\documentclass[]{unifiedreward}
\usepackage{fix-cm}
\usepackage{helvet}           

\usepackage{nicefrac}         
\usepackage{siunitx}         

\usepackage{longtable}
\usepackage{booktabs}

\usepackage{tabularx}         
\usepackage{array}           
\usepackage{makecell}         
\usepackage{threeparttable}  
\usepackage{xspace}

\usepackage{wrapfig}          
\usepackage{float}            
\usepackage[export]{adjustbox}

\usepackage{tikz}             
\usepackage{pgfplots}         
\usepackage{pgf-pie}

\usepackage{listings}        
\usepackage{ragged2e}        
\usepackage{comment}

\usepackage{pifont}           
\usepackage{fontawesome}

\usepackage{enumitem}        
\usepackage{titletoc}         
\usepackage{minitoc}          
\usepackage[toc,page,header]{appendix}

\usepackage{url}              
\usepackage[hang,flushmargin]{footmisc}  
\pgfplotsset{compat=1.18}
\definecolor{lightblue}{RGB}{200, 230, 255}  
\definecolor{headerblue}{RGB}{150, 200, 255}

\newcounter{examplebox}

\makeatletter
\newcommand\blfootnote[1]{%
  \begingroup
  \renewcommand\thefootnote{}\footnote{#1}%
  \addtocounter{footnote}{-1}%
  \endgroup
}
\makeatother

\title{
    WorldReward: Reward Modeling for Camera-Conditioned World Models
}

\author{
    Yibin Wang\textsuperscript{1,2,3},
    Zehan Wang\textsuperscript{2$^\ddagger$},
    Junshu Tang\textsuperscript{2},
    Zhimin Li\textsuperscript{2},
    Yujie Zhou\textsuperscript{4},
    Jiazi Bu\textsuperscript{4},
    Pengyang Ling\textsuperscript{6},
    Feng Han\textsuperscript{1,3},
    Zhixiong Zhang\textsuperscript{3,4},
    Long Xing\textsuperscript{5},
    Shengyuan Ding\textsuperscript{1,3}, \\
    Ziang Li\textsuperscript{3},
    Cheng Jin\textsuperscript{1,3$^\dagger$},
    Yuhang Zang\textsuperscript{5$^\dagger$},
    Jiaqi Wang\textsuperscript{3$^\dagger$},
    Tianyu Pang\textsuperscript{2$^\dagger$}
}

\affiliation[1]{\mbox{Fudan University}}
\affiliation[2]{\mbox{Tencent Hunyuan}}
\affiliation[3]{\mbox{Shanghai Innovation Institute}}
\affiliation[4]{\mbox{Shanghai Jiao Tong University}}
\affiliation[5]{\mbox{Shanghai Artificial Intelligence Laboratory}}
\affiliation[6]{\mbox{Independent Researcher}}

\newcommand{\ourmethod}{{\textsc {WorldReward}}\xspace}
\newcommand{\ourbench}{{\textsc {WorldReward-Bench}}\xspace}

\abstract{Camera-conditioned world models generate interactive videos in which commanded actions should induce the expected scene changes while appearance, geometry, and temporal dynamics remain coherent.
Existing rewards typically assess these requirements separately: geometry-based rewards estimate trajectory execution but cannot judge the visual quality of the executed motion, whereas image-based rewards measure frame quality without capturing action execution or temporal dynamics.
We posit that a vision-language model (VLM) offers a shared reasoning space for relating actions to their visual outcomes.
However, judging a complete long video against its full action sequence creates a lengthy, noisy context in which short-lived local action evidence can be missed or diluted.
To address these challenges, we present \ourmethod, a VLM-based pairwise preference reward model that unifies action-consistency and visual-quality evaluation for camera-conditioned world models.
\ourmethod decomposes paired videos into action-aligned chunks and organizes each chunk into structured visual evidence, enabling the model to evaluate the execution of each action together with visual quality. 
Chunk-level decisions are then aggregated by voting into separate video-level action and visual-quality preferences.
To train \ourmethod, we construct a large-scale reasoning-augmented preference dataset using structured judgments generated by a frontier VLM and refined through multi-turn tool-based agent auditing and targeted human review.
We further introduce \ourbench, a human-annotated benchmark that measures reward-model agreement with human preferences across action consistency, appearance quality, and motion quality.
On \ourbench, \ourmethod achieves the highest agreement on all three dimensions, exceeding GPT-5.5 by 3.42, 1.45, and 3.56 percentage points on action, appearance, and motion, respectively.
When used for reinforcement learning (RL) post-training of HY-WorldPlay~1.5, it consistently improves both action execution and visual quality across short- to long-term horizons.

}

\checkdata[Website]{\url{https://codegoat24.github.io/WorldReward}}

\begin{document}
\maketitle
\blfootnote{$^\dagger$Corresponding authors. $^\ddagger$Project lead.}

\section{Introduction}

Video-based world models simulate how a visual environment evolves in response to user controls, progressing from future-observation prediction under latent or discrete interactions~\cite{genie,gamengen,nwm} to explicit camera-trajectory control~\cite{cameractrl,cameractrl2} and real-time, long-horizon interactive generation driven by keyboard or mouse inputs~\cite{gamegenx,gamecraft,matrixgame2,worldplay}.
A generated video is useful only if it faithfully executes the commanded controls while keeping geometry, appearance, and temporal dynamics coherent over the whole horizon.
Reward models that measure both properties are therefore central to this setting: they determine how world models are evaluated and, increasingly, how they are optimized, as reinforcement learning (RL) post-training with suitable rewards improves camera control~\cite{worldcompass}.

Reward modeling for camera-conditioned world models raises three challenges. \textbf{1) Coupled requirements.} The same visual change can indicate correct or incorrect execution depending on the commanded motion, and two videos with similar trajectory accuracy can still differ in appearance or dynamics, so action consistency and visual quality should be judged from a shared interpretation of the video rather than as separate outcomes. \textbf{2) Localized evidence.} A forward or turning command manifests within a few frames of a long sequence, so a judge must locate this short-lived transition without being overwhelmed by the full video. \textbf{3) Long-horizon attribution.} Failures accumulate as generation proceeds, so local motion errors or visual degradation must be attributable to the action segment that produced them, and the resulting judgments must yield separate action and visual-quality preferences that can serve as RL signals.

Existing rewards fall short on these requirements. \textit{Geometry-based rewards} recover the camera trajectory from generated frames with 3D foundation models and compare it with the commanded actions~\cite{depthanything3,worldmirror}. They measure geometric trajectory consistency but ignore the visual quality of the executed motion, such as temporal stability, dynamic plausibility, and generation artifacts. \textit{Image-based rewards} such as HPSv3~\cite{ma2025hpsv3} score sampled frames independently, so a visually appealing frame-level score can still overlook flickering, motion discontinuities, appearance drift, or inconsistent dynamics. \textit{Combined rewards} in WorldCompass~\cite{worldcompass} pair these two signals for RL post-training, which is effective but leaves action execution and visual quality assessed by heterogeneous, decoupled systems. \textit{General video preference models}~\cite{videoalign,unifiedrewardflex,unifiedreward_think} capture perceptual quality but do not verify whether the commanded actions are followed. \textit{Direct VLM judging} over all frames and the complete action sequence creates a long, noisy multimodal context: sparse frame sampling misses short-lived transitions, dense sampling inflates the context further, and judging the video as a whole lets local failures be diluted by the overall impression.

We propose \ourmethod, a VLM-based pairwise preference reward model that grounds unified action-consistency and visual-quality evaluation in localized action--video evidence. Unlike prior rewards that score trajectory execution and frame quality with separate systems, \ourmethod derives both preferences from a single model reasoning over the same evidence. Unlike direct whole-video VLM judging, it follows a local-to-global scheme: each long video pair is divided into temporally aligned chunks of four consecutive actions, each chunk is judged from structured visual evidence, and chunk-level decisions are aggregated by voting into separate global action and visual-quality preferences (Figure~\ref{fig:reason-pipeline}).

This design follows from how action execution manifests visually. Camera actions leave direction-specific evidence: when the camera moves forward, visible content should gradually enlarge; when it tilts upward, existing content should shift downward as new content enters from the top. Such evidence is best verified by comparing a few frames around each action, which is what the chunk input provides. The source image and caption anchor scene identity, a paired frame-grid shows the start, middle, and end of each action, and action-level panels highlight each first-to-last frame transition. Within this compact context, a VLM can check each action against its commanded direction while also examining temporal consistency, dynamic generation quality, and artifact/structure integrity, yielding action and visual-quality decisions from one interpretation of the same frames. Voting over chunks then prevents a single strong or weak segment from dominating the video-level preference. The ablations in Table~\ref{tab:ablation-model-design} support each component: removing the source image, the frame grid, or the action-level panels lowers agreement with human preferences, and structured reasoning supervision improves it further.

Training such a model requires reasoning-augmented supervision at scale, and evaluating it requires human preferences along separate dimensions. We construct a preference dataset through the pipeline in Figure~\ref{fig:train-data-pipeline}: paired outputs from multiple world models~\cite{worldplay,lingbotworld,infiniteworld,yume15,sanawm,matrixgame2,matrixgame3} under matched conditions, chunk-level reasoning distilled from Gemini~3.1 Pro~\cite{gemini31pro}, multi-turn auditing by a tool-using agent based on GPT-5.5~\cite{gpt55systemcard}, and targeted human calibration of agent-revised samples. Table~\ref{tab:ablation-annotation-qc} shows that agent auditing provides most of the gain over direct distillation and that human calibration adds a further improvement. We also introduce \ourbench, a human-annotated benchmark of 760 paired generations that share the same source image, caption, and trajectory, covering diverse trajectory families, visual styles, and world-model sources (Figure~\ref{fig:benchmark}), with independent labels for action consistency, appearance quality, and motion quality.


On \ourbench, \ourmethod achieves the highest agreement with human preferences on all three dimensions, outperforming proprietary VLM judges (GPT-5.5~\cite{gpt55systemcard} and Gemini~3.1 Pro~\cite{gemini31pro}), visual preference models (e.g., HPSv3~\cite{ma2025hpsv3}), and geometric trajectory estimators (e.g., DepthAnything3~\cite{depthanything3}). Although its supervision is distilled from the two proprietary VLMs, it surpasses both after annotation refinement. Used as the reward for clip-level RL post-training of HY-WorldPlay~1.5~\cite{worldplay}, it improves both action execution and visual quality over the base model and over WorldCompass across short- to long-term horizons, and the gains are corroborated by GPT-5.5 and human evaluators under the same pairwise protocol.



Our contributions are summarized as follows:
\textbf{1)} \textbf{Unified reward model.} We propose \ourmethod, to our knowledge the first VLM-based pairwise reward model that unifies action-consistency and visual-quality evaluation for camera-conditioned world models, built on a chunk-level reasoning paradigm that judges structured action-aligned evidence and aggregates chunk decisions by voting into separate global preferences.
\textbf{2)} \textbf{Reasoning-augmented preference data.} We construct a large-scale reasoning-augmented preference dataset through frontier-VLM distillation, multi-turn tool-based agent auditing, and targeted human calibration, and show that this annotation refinement is the main source of the reward model's advantage over direct VLM judging.
\textbf{3)} \textbf{Human-annotated benchmark.} We introduce \ourbench, a human-annotated benchmark of 760 paired camera-conditioned generations with independent labels for action consistency, appearance quality, and motion quality.
\textbf{4)} \textbf{Empirical gains.} \ourmethod outperforms open-source and proprietary reward baselines on \ourbench, and its action and visual-quality preferences serve as effective reward signals for RL post-training of HY-WorldPlay~1.5, improving both action execution and visual quality across generation horizons.
\section{Related Work}

\paragraph{Camera-conditioned world models.}
Video-based world models predict future observations under learned or explicit user controls~\cite{genie,gamengen,nwm}. Camera-controlled video generation further introduces continuous trajectory conditioning for viewpoint manipulation and scene exploration~\cite{cameractrl,cameractrl2}. Interactive game and open-world models extend this paradigm with keyboard or mouse controls~\cite{gamegenx}, real-time autoregressive streaming~\cite{matrixgame2,worldplay,lingbotworld}, and long-range history conditioning~\cite{gamecraft,matrixgame3,infiniteworld}. These systems couple action controllability with geometric, appearance, and temporal quality, motivating reward signals that assess both commanded motion and its visual realization.

\paragraph{RL post-training for visual generation.}
RL and preference optimization have been widely used to align image and video diffusion or flow models, through policy-gradient fine-tuning~\cite{ddpo,dpok}, reward backpropagation~\cite{alignprop}, direct preference optimization~\cite{diffusiondpo}, human-feedback video alignment~\cite{LiFT,videoalign}, online reinforcement on the forward process~\cite{diffusionnft}, and hybrid-policy self-distillation~\cite{bu2026hpsd}. Within flow-model RL, GRPO variants explore group-relative policy updates~\cite{flowgrpo}, pairwise preference rewards~\cite{prefgrpo}, fine-grained preference alignment~\cite{finegrpo}, augmented condition views~\cite{mvgrpo}, denser temporal credit assignment~\cite{pavegrpo}, and capability-aware sampling and advantage estimation~\cite{adagrpo}. We build on this line by adopting the clip-level DiffusionNFT protocol of WorldCompass~\cite{worldcompass,diffusionnft} and the pairwise win-rate formulation of Pref-GRPO~\cite{prefgrpo}; our contribution lies in the reward signals rather than the optimization algorithm.

\paragraph{Visual preference and reward models.}
Visual reward models broadly follow two paradigms. Discriminative reward models learn a scalar scoring function from human preference data, providing efficient ranking signals for generated images~\cite{pickscore,hpsv2,ma2025hpsv3}. Generative reward models instead adopt VLMs as judges that compare candidates and produce multi-aspect evaluation reasoning~\cite{unifiedreward,unifiedreward_think,unifiedrewardflex,videoscore2}, and such judges can be further reinforced with agentic tool use and visual reasoning~\cite{armthinker}. Video-oriented reward learning extends preference modeling to temporally structured video quality~\cite{LiFT,videoalign,gtsvj}. These models provide strong general-purpose feedback for visual generation, but they do not condition on a commanded action. \ourmethod follows the VLM-as-judge paradigm of UnifiedReward~\cite{unifiedreward,unifiedreward_think,unifiedrewardflex} and extends it to camera-conditioned world models, where each judgment must additionally verify the local visual transition produced by a commanded action.

\paragraph{Reward signals for world models.}
Geometry foundation models recover camera trajectories and scene structure from image sequences~\cite{dust3r,vggt,depthanything3,worldmirror}, and their estimates support action-following rewards that compare the recovered motion with the commanded trajectory~\cite{worldcompass}. Related signals derive verifiable rewards from inverse dynamics~\cite{rlir} or from geometric and perceptual consistency~\cite{grndctrl}. These signals characterize the geometric execution of camera actions but do not assess the visual quality of the resulting motion, so WorldCompass~\cite{worldcompass} pairs a geometry reward with an image-based HPSv3 reward~\cite{ma2025hpsv3}, leaving the two aspects to heterogeneous systems. For embodied world models, ReWorld~\cite{reworld} trains a hierarchical multi-dimensional reward model covering physical realism, task completion, embodiment plausibility, and visual quality, and Reward as an Agent~\cite{rewardagent} evaluates generated behaviors with an agentic reward to mitigate reward hacking. These works target task-oriented embodied generation, whereas \ourmethod targets camera-conditioned generation, grounds each judgment in a local action--video chunk, and derives action-consistency and visual-quality preferences from a shared VLM analysis.
\section{Method}

\subsection{Problem Formulation}

Camera-conditioned world models aim to generate videos whose scene evolution follows a prescribed camera/action trajectory while preserving the visual content and dynamic plausibility of the source scene. A reliable reward for this setting should therefore evaluate two coupled aspects: whether the generated scene changes execute the commanded actions, and whether the resulting video remains visually faithful, temporally coherent, and free of severe structural artifacts. We study this reward modeling problem under a paired comparison setting. Given an input visual condition consisting of a source image $x_0$ and its caption $d$, together with a camera/action trajectory $a_{1:N}$ of $N$ action steps, two candidate videos $V^A$ and $V^B$ are generated under the same condition and trajectory. The reward model predicts a dimension-specific preference for action consistency or visual quality:
\begin{equation}
    p_\theta^{m}\left(A \succ B \mid x_0, d, a_{1:N}, V^A, V^B\right),
    \quad m \in \{\mathrm{act}, \mathrm{vis}\}.
\end{equation}
Rather than presenting the full video as a single unstructured input, \ourmethod evaluates a sequence of chunk-level comparison inputs:
\begin{equation}
    z_k = \left(d, a_{s_k:e_k}, \mathcal{I}_k\right), \quad k=1,\dots,K,
\end{equation}
where $a_{s_k:e_k}$ is a local action segment and $\mathcal{I}_k$ is the structured visual input for the chunk. As shown in Figure~\ref{fig:reason-pipeline}, $\mathcal{I}_k$ is constructed from the source image $x_0$, a paired frame-grid overview, and action-level comparison panels from the decoded frame blocks in $V^A$ and $V^B$ that correspond to $a_{s_k:e_k}$. For each chunk, \ourmethod predicts action and visual-quality preferences. The action preference is obtained from action-wise comparisons within the chunk, while the visual-quality preference is obtained by jointly considering temporal consistency, dynamic generation quality, and artifact/structure integrity. The chunk-level preferences are then aggregated across $K$ chunks to produce global action and global visual-quality preferences.

\subsection{Action-Video Chunk Construction}

Directly feeding the full video pair and the complete action trajectory to a VLM introduces several difficulties: (1) the multimodal context becomes overly long because the model must process many frames from both candidates; (2) the model must track all actions at once, weakening the association between a local action and the visual evidence that verifies it; and (3) local motion errors or visual artifacts may be diluted by the global video impression and become hard to attribute to a specific action segment. We therefore decompose the action trajectory into $K$ temporally ordered chunks, where each chunk contains a short segment of consecutive actions and the corresponding decoded frame blocks from both candidates. We prepend an idle slot associated with the source frame $x_0$ and partition the resulting sequence into fixed-size chunks of four slots. For the $k$-th chunk, the reward input contains $x_0$, the image caption, the local action segment, and the temporally aligned visual evidence from $V^A$ and $V^B$.

\begin{figure}[t]
    \centering
    \includegraphics[width=\linewidth]{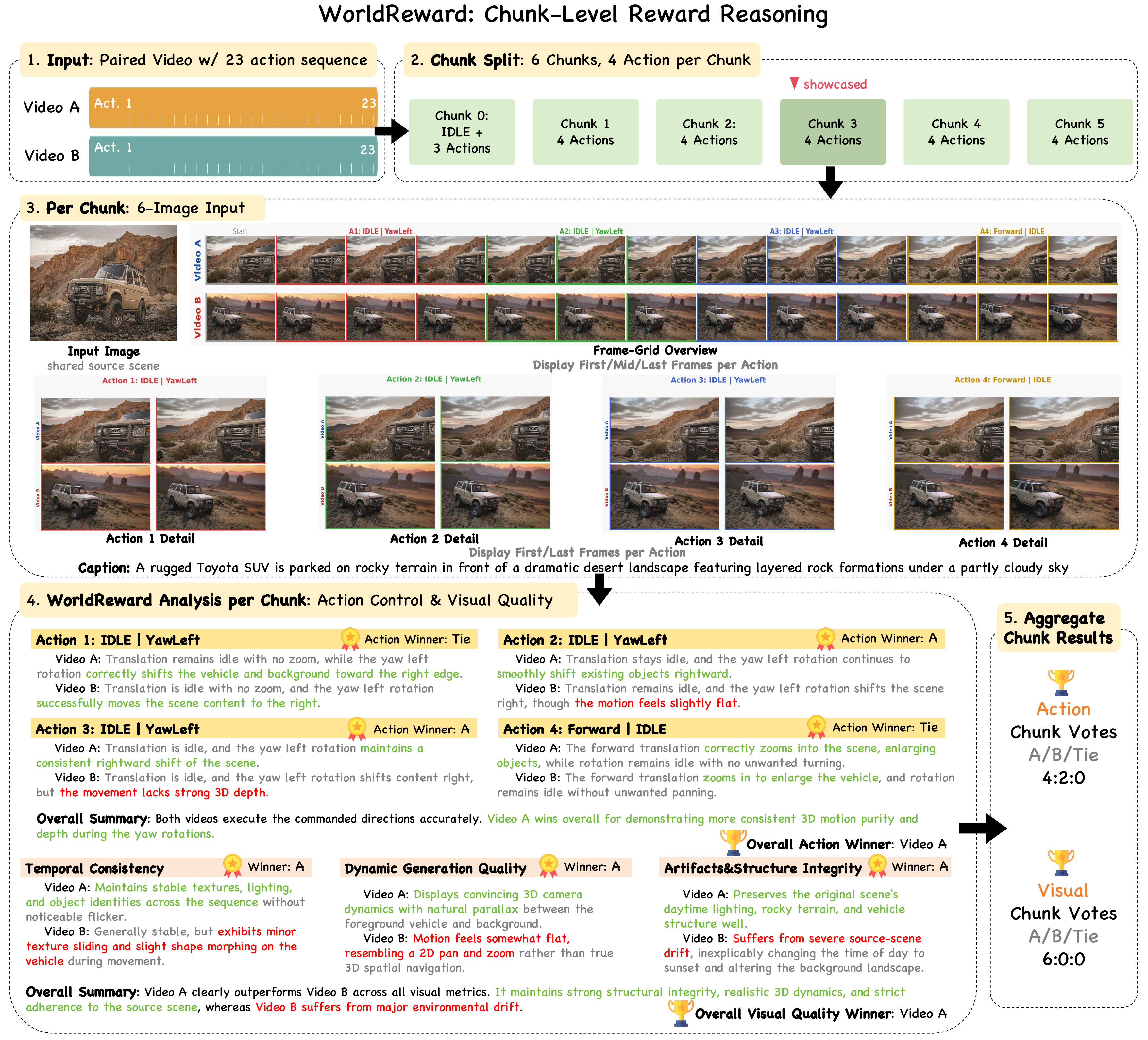}
    \caption{\textbf{Chunk-level reward reasoning in \ourmethod.} A paired video is split into action-aligned chunks. Each chunk is judged from a six-image input (source image, frame-grid overview, and four action-level panels), yielding action and visual-quality winners in $\{A,B,\mathrm{Tie}\}$ that are aggregated by voting into video-level preferences.}
    \label{fig:reason-pipeline}
\end{figure}

Figure~\ref{fig:reason-pipeline} illustrates this chunk-level input design. For each selected chunk, we build a compact multi-image input consisting of the source image, a frame-grid overview of the paired videos, and action-level detail panels. The source image provides the reference scene for detecting source-scene drift and identity changes. The frame-grid overview displays the start, middle, and end frames of each action in the chunk, allowing the model to inspect temporal consistency and the strength of generated dynamics across the chunk. Each action-level panel further compares the first and last frames of the corresponding action segment from the two videos, making the local scene transition easier to judge. The caption is provided together with these visual inputs to preserve semantic context. This organization allows the model to inspect the scene changes induced by each local action segment before judging the full video, making short-lived failures easier to identify, including incorrect camera movement, abrupt geometry changes, appearance drift, flickering, and motion discontinuities. The structured-evidence ablations in Table~\ref{tab:ablation-model-design} support this design: removing the source image, the frame-grid overview, or the action-level panels consistently lowers agreement with human preferences.

\subsection{Chunk-Level Reward Reasoning}

For each action-video chunk, \ourmethod performs pairwise reward reasoning at both action and visual-quality levels. For action control, the model first compares the two videos for each action in the chunk, judging whether the local scene transition follows the commanded camera/action direction. It then summarizes the action-wise decisions into an overall action winner for the chunk. For visual quality, the model evaluates three complementary aspects: temporal consistency, which checks stability across frames; dynamic generation quality, which checks whether the observed motion reflects plausible camera-conditioned 3D dynamics; and artifact/structure integrity, which checks visual artifacts, structural degradation, and source-scene preservation. Through this process, the model produces chunk-level action and visual-quality preferences, each supported by the reasoning over the corresponding criteria. The reasoning-supervision ablations in Table~\ref{tab:ablation-model-design} support this hierarchical design: adding the overall comparison summary and the per-video analysis to preference-only supervision progressively improves agreement with human preferences.

Let $r^{\mathrm{act}}_k$ and $r^{\mathrm{vis}}_k$ denote the action and visual-quality winners predicted for chunk $k$, where each winner belongs to $\{A,B,\mathrm{Tie}\}$. For dimension $m$, let $n_c^m=\sum_{k=1}^{K}\mathbf{1}[r_k^m=c]$ be the number of chunks favoring candidate $c\in\{A,B\}$. We aggregate the chunk decisions by voting:
\begin{equation}
    R^m =
    \begin{cases}
        A, & n_A^m > n_B^m,\\
        B, & n_B^m > n_A^m,\\
        \mathrm{Tie}, & n_A^m = n_B^m,
    \end{cases}
    \quad m \in \{\mathrm{act}, \mathrm{vis}\}.
    \label{eq:chunk-voting}
\end{equation}
Thus, chunks predicted as Tie do not favor either candidate, and equal numbers of votes for A and B produce a global Tie. This yields the global action winner $R^{\mathrm{act}}$ and the global visual-quality winner $R^{\mathrm{vis}}$ for the complete video pair. The voting formulation follows a local-to-global principle: the model first grounds its decision in temporally localized evidence, and then combines the fine-grained judgments into video-level reward signals.

\subsection{Reasoning-Augmented Preference Data}
\label{sec:preference-data}

\begin{figure}[t]
    \centering
    \includegraphics[width=\linewidth]{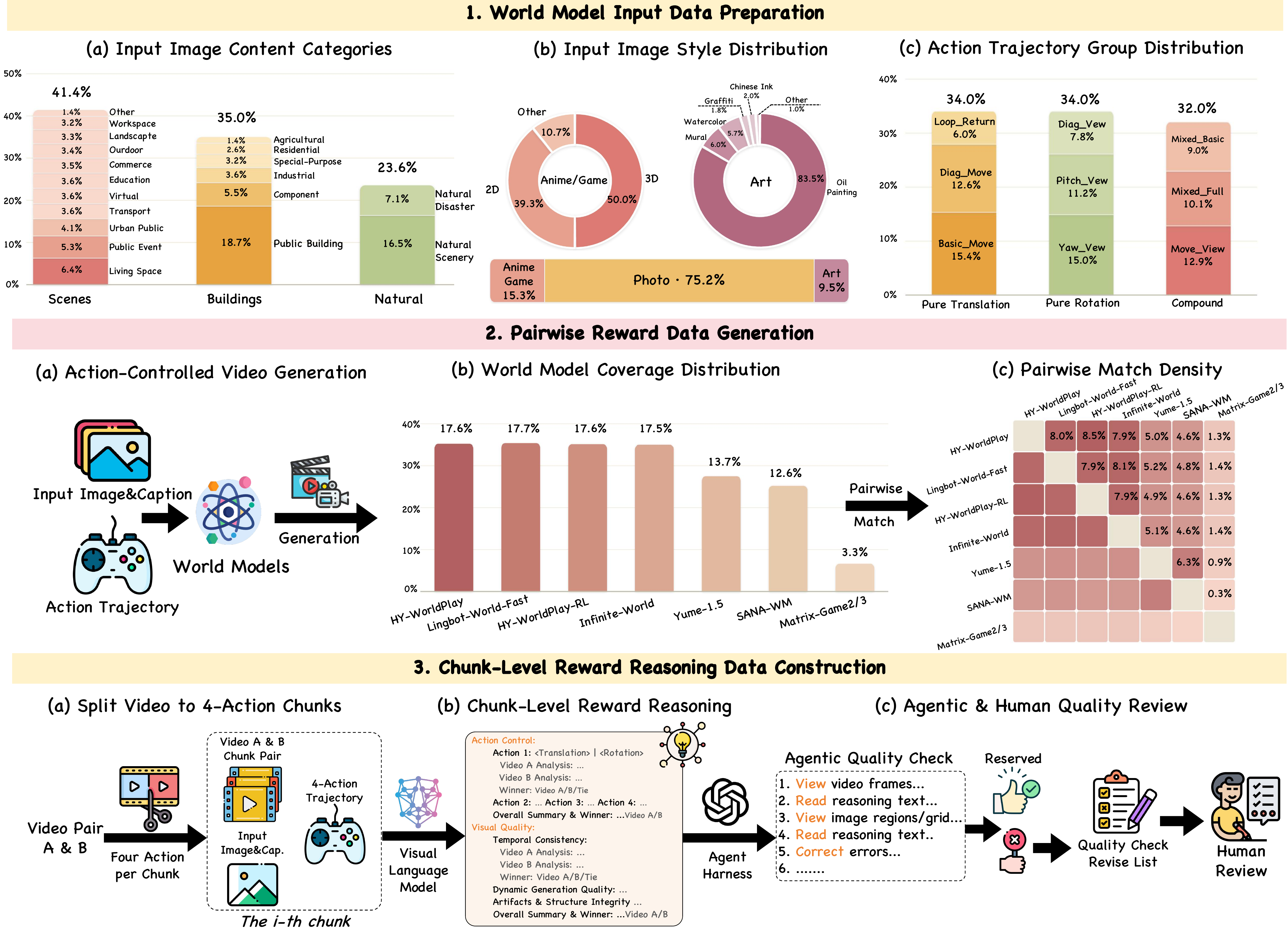}
    \caption{\textbf{Reasoning-augmented preference-data pipeline.} (1) Input preparation over image content, style, and trajectory groups. (2) Pairwise generation from eight world-model variants, with per-model coverage and pair-assignment density. (3) Chunk-level reasoning annotation by a VLM, followed by agent auditing and human review.}
    \label{fig:train-data-pipeline}
\end{figure}

Training a reliable reward model requires supervision that spans diverse visual generation distributions and action trajectories, while capturing temporally localized failures across a wide range of world-model outputs. We therefore construct a reasoning-augmented preference dataset through the three-stage pipeline shown in Figure~\ref{fig:train-data-pipeline}: preparing diverse world-model inputs, generating paired world-model outputs, and constructing chunk-level reasoning annotations with agent-assisted quality control and human review.

\paragraph{World-model input data preparation.}
We first prepare the shared generation conditions for camera-conditioned world models. Each condition is a tuple containing an input image, its caption, and an action trajectory. For each model pair, the same condition is used to generate both videos, ensuring matched visual content and camera controls.

\begin{sloppypar}
For image coverage, we organize input images along two axes, as shown in Figure~\ref{fig:train-data-pipeline}. The content categories include scene-centric images, built environments, and natural scenes, with fine-grained categories such as living spaces, public events, urban public areas, transport, public buildings, industrial or residential scenes, natural scenery, and natural disasters. Styles are primarily photorealistic, with additional anime, game, and artistic images such as oil paintings, watercolors, murals, graffiti, and ink-style content. This coverage is intended to expose reward models to action and quality failures across both realistic and stylized world-model outputs.
\end{sloppypar}

For action coverage, we organize trajectories into three families:
\begin{itemize}[leftmargin=*,topsep=2pt,itemsep=1pt,parsep=0pt]
    \item \textbf{Pure Translation} contains position-only camera motions without rotation. \textit{Basic\_Move} covers single-axis translations, including forward, backward, left, and right. \textit{Diag\_Move} combines two translation axes at the same time, such as forward+left or backward+right. \textit{Loop\_Return} moves away from the starting position and then returns along a symmetric reverse path, forming a closed go-and-return trajectory.
    \item \textbf{Pure Rotation} contains viewpoint-only camera motions with fixed camera position. \textit{Yaw\_View} rotates the camera horizontally to the left or right while keeping pitch unchanged. \textit{Pitch\_View} rotates the camera upward or downward while keeping yaw unchanged. \textit{Diag\_View} combines yaw and pitch changes simultaneously, such as yaw-left with pitch-up.
    \item \textbf{Compound} contains trajectories that combine translation and rotation. \textit{Mixed\_Basic} executes a translation segment and a rotation segment sequentially, so the two basic motions are not simultaneous. \textit{Mixed\_Full} applies one translation axis and one rotation axis at the same time, such as forward motion with yaw-left rotation. \textit{Move\_View} is a stronger simultaneous combination involving multi-axis translation and multi-axis rotation, such as forward-right motion with yaw-left and pitch-up rotation.
\end{itemize}
This taxonomy progresses from atomic to increasingly coupled camera control. Basic and diagonal trajectories test directional execution, loop-return trajectories expose accumulated drift and failure to recover the source viewpoint, and compound trajectories stress whether translation and rotation can be executed sequentially or simultaneously without destabilizing scene geometry.

\paragraph{Pairwise reward data generation.}
Given each input image-caption pair and action trajectory, we generate action-controlled videos using a diverse set of camera-conditioned world models. As illustrated in Figure~\ref{fig:train-data-pipeline}, our generation pool includes HY-WorldPlay and its RL-post-trained variant~\cite{worldplay,worldcompass}, LingBot-World-Fast~\cite{lingbotworld}, Infinite-World~\cite{infiniteworld}, Yume-1.5~\cite{yume15}, SANA-WM~\cite{sanawm}, and Matrix-Game models~\cite{matrixgame2,matrixgame3}. These eight model variants differ in architecture, training data, interaction horizon, and controllability, producing a broad range of action-following errors and visual-quality failures.

Both models in a pair receive the same source image, caption, and action trajectory, controlling for visual content and action difficulty when deriving the preference. Pair assignments are balanced across the major full-control model combinations. We additionally apply compatibility-aware assignment so that a model is evaluated only on controls supported by its interface. In particular, the Matrix-Game models support only a subset of the trajectory families, so their model-pair frequencies are lower than those of the four broadly compatible models. This filtering prevents unsupported controls from inducing trivial preferences, while balanced assignment reduces bias toward particular model identities.

\paragraph{Chunk-level reward reasoning data construction.}
For each matched video pair, we divide the idle-augmented trajectory into fixed-size chunks and align each chunk with the corresponding frames in both videos. We then sample chunks for reasoning annotation and reward-model training. The chunk-level reasoning example in Figure~\ref{fig:reason-pipeline} also illustrates the content of both the distilled annotations and the final training samples, including the structured visual evidence, caption, local action sequence, action-control and visual-quality reasoning, and their corresponding preferences.

First, we distill structured reward reasoning from Gemini 3.1 Pro~\cite{gemini31pro} using the chunk-level protocol above. The resulting annotations pair action and visual-quality preferences with evidence-based rationales, providing richer supervision than preference labels alone.

Second, we introduce an agent-assisted quality-control stage to reduce distillation noise. Although an API-based critic can also verify an existing annotation, it typically receives a fixed set of images in one request and must complete the inspection within a single response. We therefore adopt an agent-harness-based QC procedure with GPT-5.5~\cite{gpt55systemcard}, which adaptively invokes image-reading tools over multiple rounds. Rather than loading every detailed panel into the initial context, the agent first inspects the source image and frame-grid overview, uses its observations to select the next action-level panel, and revisits previous evidence when cross-checking is needed. This adaptive process keeps each inspection focused on the disputed chunk and prevents its local evidence from being diluted by unrelated frames.

As summarized in Algorithm~1, the agent either retains the original verdict or returns a diff that localizes each proposed correction to a chunk and evaluation dimension, making the revision easy to trace and review.

\begin{tcolorbox}[
    colback=gray!3,
    colframe=gray!45,
    title=\textbf{Algorithm 1: Agent-Harness Quality Control},
    fonttitle=\small,
    fontupper=\small,
    left=4pt,
    right=4pt,
    top=4pt,
    bottom=4pt
]
\textbf{Input:} Gemini verdict $y$, caption $d$, local action sequence $a$, and visual evidence $\mathcal{E}=\{x_0, \texttt{frame\_grid}, \texttt{action images}\}$.\\
\textbf{Output:} quality-controlled verdict $\hat{y}$.
\begin{enumerate}[leftmargin=*,topsep=2pt,itemsep=1pt,parsep=0pt]
    \item Initialize $\hat{y}\leftarrow y$ and an inspection history $\mathcal{H}\leftarrow\emptyset$.
    \item Invoke image-reading tools to inspect $x_0$ and the frame-grid overview; append the observations to $\mathcal{H}$.
    \item For inspection rounds $t=1,\ldots,T$:
    \begin{enumerate}[label=(\alph*),leftmargin=*,topsep=1pt,itemsep=0pt,parsep=0pt]
        \item Audit the current verdict against $\mathcal{H}$ and identify chunk--dimension pairs that remain uncertain or inconsistent with the observed evidence.
        \item If no further evidence is required, terminate the inspection loop.
        \item Select the relevant action-level panel, invoke the image-reading tool, and append the new observation to $\mathcal{H}$. Previously inspected panels may be revisited when cross-checking is needed.
    \end{enumerate}
    \item If no inconsistency is recorded, return $\hat{y}=y$; otherwise, produce and apply a localized diff $\Delta$ from the complete inspection history to obtain $\hat{y}$.
    \item Forward agent-revised samples to human calibration.
\end{enumerate}
\end{tcolorbox}

Finally, samples revised by the agent undergo human calibration. Reviewers inspect the visual evidence alongside the proposed diff and determine whether each revision is supported, concentrating human effort on records for which the agent identifies a potential annotation error. Table~\ref{tab:qc-statistics} summarizes the outcomes of agent-assisted QC and stratified human calibration. All revision statistics refer to chunk-level reasoning annotations. Among the revised samples, 67.4\% modify only the reasoning text, 23.8\% modify only per-dimension winners, 0.9\% modify only chunk-level winners, and 7.9\% modify both per-dimension and chunk-level winners. This distribution shows that the agent improves supervision at multiple granularities, while the 87.0\% human confirmation rate supports the reliability of its proposed corrections.

\begin{table}[t]
    \centering
    \caption{\textbf{Quality-control outcomes.} Top: chunk-level annotations retained or revised by the agent harness, with a breakdown of revisions. Bottom: agent-proposed revisions confirmed or rejected by human reviewers.}
    \label{tab:qc-statistics}
    \small
    \setlength{\tabcolsep}{4pt}
    \begin{tabularx}{\linewidth}{@{}p{0.30\linewidth}Xr@{}}
        \toprule
        \textbf{Stage} & \textbf{Outcome} & \textbf{Percentage} \\
        \midrule
        Agent-assisted QC & Original retained & 57.9\% \\
        & Any revision & 42.1\% \\
        \addlinespace[2pt]
        \multicolumn{3}{@{}l}{\textit{Breakdown among agent-revised samples}} \\
        & Reasoning text only & 67.4\% \\
        & Per-dimension winner only & 23.8\% \\
        & Chunk-level winners only & 0.9\% \\
        & Both winner types & 7.9\% \\
        \midrule
        Human calibration & Revision confirmed & 87.0\% \\
        & Revision rejected & 13.0\% \\
        \bottomrule
    \end{tabularx}
\end{table}

The resulting dataset combines chunk-level multimodal inputs, evidence-based reward reasoning, and separate action-control and visual-quality preferences, providing the supervision used to train \ourmethod.

\subsection{\ourmethod Training}

\paragraph{Training.}
We formulate reward-model training as supervised fine-tuning of a multimodal large language model. Let $\mathcal{D}=\{(z_i,s_i)\}_{i=1}^{M}$ denote the chunk-level training set, where $z_i$ is the structured multimodal input defined above and $s_i=(s_{i,1},\ldots,s_{i,L_i})$ is the target response. Each response contains the action-control and visual-quality reasoning for the chunk, followed by their categorical preferences in $\{A,B,\mathrm{Tie}\}$. Using teacher forcing, we optimize the standard autoregressive language-modeling objective:
\begin{equation}
    \mathcal{L}_{\mathrm{SFT}}(\theta)
    = -\frac{1}{\sum_{i=1}^{M}L_i}
    \sum_{i=1}^{M}\sum_{t=1}^{L_i}
    \log p_{\theta}\!\left(s_{i,t}\mid z_i,s_{i,<t}\right).
    \label{eq:sft-loss}
\end{equation}
The loss is computed over the target response tokens, while the caption, local action sequence, and visual inputs serve as conditioning context. Consequently, this objective teaches the model to explain the relevant visual evidence and produce the corresponding action and visual-quality preferences.

\subsection{World Model Post-Training with \ourmethod}
\label{sec:rl-posttraining}

We further use \ourmethod to provide reinforcement signals for post-training HY-WorldPlay~1.5~\cite{worldplay}. We retain the clip-level rollout and DiffusionNFT~\cite{diffusionnft} optimization framework of WorldCompass~\cite{worldcompass}. Unlike its geometry-based action reward and image-based quality reward, which assess trajectory execution and visual quality through separate reward systems, \ourmethod derives both preferences from a shared interpretation of the commanded actions, the induced scene changes, and their visual outcomes. The two preferences remain separate optimization signals, allowing action execution and visual quality to be balanced explicitly, but they are grounded in the same action-aligned visual context and produced by a single reward model.

\paragraph{Clip-level rollout.}
Let $\pi_{\phi}$ denote the world-model policy and $\pi_{\bar\phi}$ its exponential-moving-average copy used for rollout, and let $a_n$ denote the local action segment assigned to clip $n$. At each iteration, we select a target clip index $n$, generate the shared autoregressive prefix $x_{1:n-1}$ once, and sample a group of $G$ candidate target clips from independent initial noises:
\begin{equation}
    x_{1:n-1}\sim\pi_{\bar\phi}(\cdot\mid a_{1:n-1},x_0,d),
    \qquad
    x_n^{(i)}\sim\pi_{\bar\phi}(\cdot\mid x_{1:n-1},a_n,x_0,d),
    \quad i=1,\ldots,G.
    \label{eq:rl-rollout}
\end{equation}
The target rollout clip is temporally aligned with the action-video chunk used by \ourmethod. For every unordered candidate pair $\{i,j\}$, we construct the same chunk-level reward input as described above: the original source image $x_0$ and its caption, the local action sequence, a paired frame-grid overview, and action-level comparison panels. The source image is the first frame of the complete video sequence. All candidate clips share the same prefix and begin to diverge only after its final frame, which defines their common temporal starting point.

\paragraph{Pairwise preference rewards.}
For each candidate pair, \ourmethod jointly reasons over the same action-aligned visual context and produces two dimension-specific preferences: an action winner and a visual-quality winner,
$y_{ij}^{m}\in\{i,j,\mathrm{Tie}\}$ for $m\in\{\mathrm{act},\mathrm{vis}\}$. Following the pairwise win-rate formulation of Pref-GRPO~\cite{prefgrpo} and its multi-dimensional use in UnifiedReward-Flex~\cite{unifiedrewardflex}, we convert each categorical comparison into the contribution
\begin{equation}
    \omega_i(y_{ij}^{m})=
    \begin{cases}
        1, & y_{ij}^{m}=i,\\
        \tfrac{1}{2}, & y_{ij}^{m}=\mathrm{Tie},\\
        0, & y_{ij}^{m}=j.
    \end{cases}
    \label{eq:pairwise-contribution}
\end{equation}
The reward of candidate $i$ is its normalized win rate against the other $G-1$ candidates:
\begin{equation}
    R_i^{m}=\frac{1}{G-1}\sum_{j\ne i}\omega_i(y_{ij}^{m}),
    \qquad m\in\{\mathrm{act},\mathrm{vis}\}.
    \label{eq:rl-win-rate}
\end{equation}

\paragraph{DiffusionNFT optimization.}
We standardize the two win-rate rewards independently within each rollout group:
\begin{equation}
    A_i^{m}=\frac{R_i^{m}-\mu_m}{\sigma_m+\epsilon},
    \qquad m\in\{\mathrm{act},\mathrm{vis}\},
    \label{eq:rl-advantage}
\end{equation}
where $\mu_m$ and $\sigma_m$ are the group mean and standard deviation. Following WorldCompass, the two advantages are combined into the optimality probability
\begin{equation}
    p_i=\frac{1}{2}+\frac{1}{2}\operatorname{clip}\!\left(
    \frac{\lambda A_i^{\mathrm{act}}+(1-\lambda)A_i^{\mathrm{vis}}}{Z},-1,1
    \right),
    \label{eq:rl-optimality}
\end{equation}
where $\lambda$ balances action execution and visual quality, and $Z$ controls the normalization scale. We then apply the negative-aware flow-matching objective. For a rollout latent $x_n^{(i)}$, noise $\varepsilon\sim\mathcal{N}(0,I)$, and diffusion time $t$, let $z_t^{(i)}=(1-t)x_n^{(i)}+t\varepsilon$ and $u^{(i)}=x_n^{(i)}-\varepsilon$. Denoting the implicit positive and negative velocity predictions by
\begin{equation}
    v_{\phi}^{+}=(1-\beta)v_{\bar\phi}+\beta v_{\phi},
    \qquad
    v_{\phi}^{-}=(1+\beta)v_{\bar\phi}-\beta v_{\phi},
    \label{eq:rl-velocity}
\end{equation}
where both predictions are conditioned on $z_t^{(i)}$, the shared prefix, local actions, and generation conditions, the training loss is
\begin{equation}
    \mathcal{L}_{\mathrm{RL}}(\phi)
    =\mathbb{E}_{i,t}\!\left[
        p_i\left\|v_{\phi}^{+}-u^{(i)}\right\|_2^2
        +(1-p_i)\left\|v_{\phi}^{-}-u^{(i)}\right\|_2^2
    \right].
    \label{eq:rl-loss}
\end{equation}
Overall, our reward and optimization design preserves the efficiency of clip-level long-horizon post-training and the flexibility of explicitly balancing action execution and visual quality. Meanwhile, the two dimension-specific reward signals are derived from a shared interpretation of action-induced scene changes and their visual outcomes, rather than being computed independently by heterogeneous task-specific reward models.

\subsection{\ourbench Construction}
\label{sec:bench-construction}

We further construct \ourbench, a human-annotated benchmark for measuring how reliably reward models reproduce human preferences on camera-conditioned world-model outputs. The benchmark contains 760 paired comparisons. Within each pair, the two videos are generated from the same source image, caption, and action trajectory, so the comparison holds the generation condition fixed and isolates differences in trajectory execution and generated visual content.

\begin{figure}[t]
    \centering
    \includegraphics[width=\linewidth]{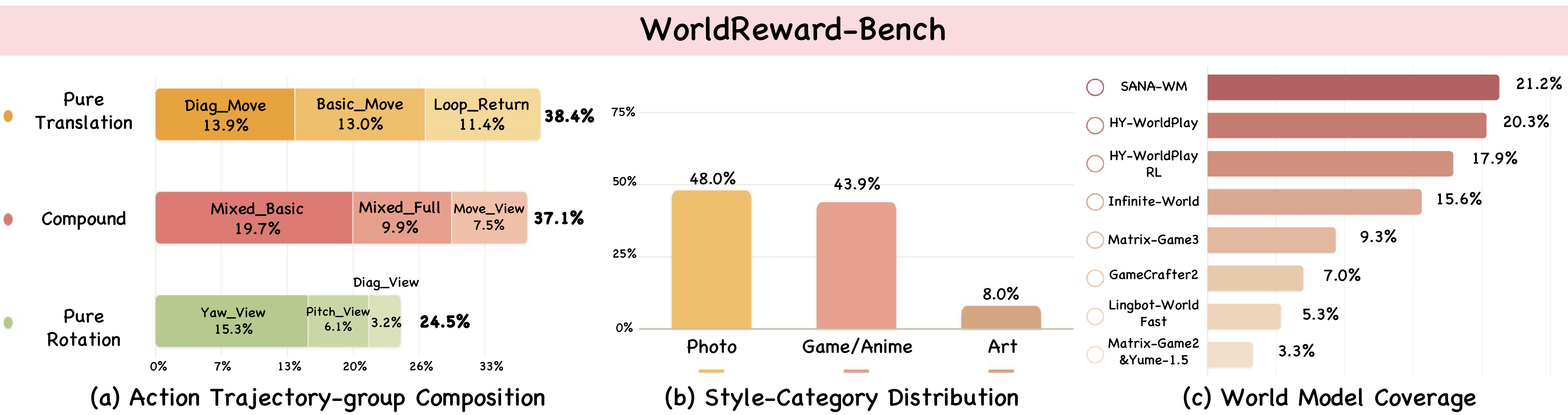}
    \caption{\textbf{Composition of \ourbench.} (a) Trajectory groups, (b) visual styles, and (c) source world models of the 760 video pairs.}
    \label{fig:benchmark}
\end{figure}

As shown in Figure~\ref{fig:benchmark}, \ourbench is designed to cover diverse camera controls, visual styles, and world-model sources. Pure translation and compound trajectories account for 38.4\% and 37.1\% of the benchmark, respectively, while pure rotation contributes the remaining 24.5\%; all trajectory sub-families introduced above are represented. The style distribution includes photorealistic scenes, game/anime content, and artistic imagery. The paired videos are drawn from a broad pool of world models~\cite{worldplay,yume15,gamecraft,matrixgame2,matrixgame3,sanawm,infiniteworld,lingbotworld}. Together, these axes expose reward models to different motion complexities, visual domains, and generator-specific failure patterns.

\paragraph{Annotation protocol.}
Each video pair is independently evaluated by three annotators. The two candidates share the same source image, caption, and action trajectory, and their model identities are hidden throughout annotation. Candidate order is randomized for each pair. Annotators assess action consistency, appearance quality, and motion quality separately, choosing from $\{A,B,\mathrm{Tie}\}$:
\begin{itemize}[leftmargin=*,topsep=2pt,itemsep=1pt,parsep=0pt]
    \item \textbf{Action consistency} measures how faithfully the scene changes follow the prescribed camera/action trajectory. Annotators examine both the direction and temporal order of the executed controls.
    \item \textbf{Appearance quality} compares visual fidelity, temporal consistency, and the preservation of scene content and structure.
    \item \textbf{Motion quality} compares the plausibility and smoothness of the generated dynamics, including discontinuous or insufficient motion.
\end{itemize}
Before annotation, annotators complete a calibration round with examples covering common camera-control errors, appearance degradation, implausible motion, and genuinely ambiguous comparisons. They select A or B only when one candidate shows a clear, observable advantage under the corresponding criterion; otherwise, they assign Tie. Each dimension is considered independently: better appearance does not imply more accurate action execution, and larger motion does not necessarily imply more plausible motion.

The benchmark label is determined by majority vote for each dimension. Cases without a majority, including three-way disagreements among A, B, and Tie, are reviewed by an additional annotator under the same criterion.

Let $y_j^m$ and $\hat{y}_j^m$ denote the human label and predicted preference for pair $j$ on dimension $m\in\{\mathrm{act},\mathrm{app},\mathrm{mot}\}$. We report three-way preference accuracy over all pairs, treating $\mathrm{Tie}$ as an explicit label:
\begin{equation}
    \mathrm{Acc}^m
    = \frac{1}{|\mathcal{B}|}\sum_{j\in\mathcal{B}}
    \mathbf{1}[\hat{y}_j^m=y_j^m].
    \label{eq:benchmark-accuracy}
\end{equation}

\FloatBarrier

\section{Experiments}
\subsection{Experimental Setup}
\begin{table}[t]
    \centering
    \caption{\textbf{Trajectory libraries.} Sub-family counts in the 224-entry training library (11 steps per trajectory) and the independently constructed 224-entry \ourbench library (23 steps).}
    \label{tab:trajectory-library}
    \small
    \setlength{\tabcolsep}{2pt}
    \begin{tabularx}{\linewidth}{@{}l*{3}{>{\centering\arraybackslash}X}@{}}
        \toprule
        \textbf{Family} &
        \multicolumn{3}{c}{\textbf{Sub-family count (Training / Benchmark)}} \\
        \cmidrule(l){2-4}
        Pure translation
        & \textit{Basic\_Move}: 24 / 30
        & \textit{Diag\_Move}: 34 / 30
        & \textit{Loop\_Return}: 12 / 18 \\
        \midrule
        Pure rotation
        & \textit{Yaw\_View}: 16 / 16
        & \textit{Pitch\_View}: 16 / 18
        & \textit{Diag\_View}: 30 / 18 \\
        \midrule
        Compound
        & \textit{Move\_View}: 68 / 36
        & \textit{Mixed\_Basic}: 14 / 28
        & \textit{Mixed\_Full}: 10 / 30 \\
        \bottomrule
    \end{tabularx}
\end{table}

\begin{table*}[t]
    \centering
    \caption{\textbf{Agreement with human preferences on \ourbench (\%).} Three-way accuracy for action consistency (Act.), appearance quality (App.), and motion quality (Mot.); Tie is an explicit label. Best and second-best are \textbf{bold} and \underline{underlined}; \texttt{--} marks a dimension the predictor does not model.}
    \label{tab:benchmark-predictor-comparison}
    \scriptsize
    \setlength{\tabcolsep}{2.0pt}
    \renewcommand{\arraystretch}{1.05}
    \resizebox{\linewidth}{!}{%
    \begin{tabular}{l*{21}{c}}
        \toprule
        \multirow{3}{*}{Reward Model}
        & \multicolumn{3}{c}{Overall}
        & \multicolumn{9}{c}{Trajectory group}
        & \multicolumn{9}{c}{Style bucket} \\
        \cmidrule(lr){2-4}\cmidrule(lr){5-13}\cmidrule(lr){14-22}
        & \multicolumn{3}{c}{All pairs}
        & \multicolumn{3}{c}{Translation}
        & \multicolumn{3}{c}{Rotation}
        & \multicolumn{3}{c}{Compound}
        & \multicolumn{3}{c}{Photo}
        & \multicolumn{3}{c}{Game/Anime}
        & \multicolumn{3}{c}{Art} \\
        \cmidrule(lr){2-4}\cmidrule(lr){5-7}\cmidrule(lr){8-10}\cmidrule(lr){11-13}\cmidrule(lr){14-16}\cmidrule(lr){17-19}\cmidrule(lr){20-22}
        & Act. & App. & Mot. & Act. & App. & Mot. & Act. & App. & Mot. & Act. & App. & Mot. & Act. & App. & Mot. & Act. & App. & Mot. & Act. & App. & Mot. \\
        \midrule
        \rowcolor{gray!10}\multicolumn{22}{l}{\textit{Closed-source VLM}} \\
        Gemini 3.1 Pro & 65.79 & \underline{80.13} & 60.79 & 64.73 & \textbf{82.19} & 64.38 & 62.90 & 83.87 & 57.53 & 68.79 & \underline{75.53} & 59.22 & 65.48 & \textbf{82.74} & 63.84 & 65.27 & 79.34 & 57.49 & 70.49 & 68.85 & 60.66 \\
        GPT-5.5 & \underline{74.21} & 79.87 & \underline{69.47} & \underline{72.95} & \underline{81.16} & \underline{74.66} & \underline{72.04} & \underline{84.41} & 62.90 & \underline{76.95} & \underline{75.53} & \underline{68.44} & \underline{76.44} & \underline{81.64} & \underline{68.77} & 71.56 & \underline{79.64} & \underline{68.86} & \textbf{75.41} & \underline{70.49} & \textbf{77.05} \\
        \midrule
        \rowcolor{gray!10}\multicolumn{22}{l}{\textit{Image/video quality reward models}} \\
        VideoAlign & -- & 61.32 & 40.13 & -- & 66.44 & 31.51 & -- & 60.22 & 45.16 & -- & 56.74 & 45.74 & -- & 61.10 & 41.10 & -- & 61.08 & 41.32 & -- & 63.93 & 27.87 \\
        UR-Flex & -- & 64.34 & 49.32 & -- & 63.18 & 47.65 & -- & 72.63 & 55.31 & -- & 60.14 & 47.10 & -- & 65.08 & 55.03 & -- & 63.32 & 46.08 & -- & 65.45 & 30.91 \\
        UR-Think & -- & 66.09 & 38.79 & -- & 65.41 & 32.88 & -- & 69.73 & 52.43 & -- & 64.41 & 35.94 & -- & 69.51 & 41.21 & -- & 64.37 & 37.43 & -- & 55.00 & 31.67 \\
        Aesthetic & -- & 69.87 & -- & -- & 66.10 & -- & -- & 72.04 & -- & -- & 72.34 & -- & -- & 66.58 & -- & -- & 75.15 & -- & -- & 60.66 & -- \\
        HPSv3 & -- & 73.68 & -- & -- & 74.66 & -- & -- & 76.34 & -- & -- & 70.92 & -- & -- & 73.15 & -- & -- & 75.45 & -- & -- & 67.21 & -- \\
        \midrule
        \rowcolor{gray!10}\multicolumn{22}{l}{\textit{Geometry estimation models}} \\
        DAv3 & 70.53 & -- & -- & 67.47 & -- & -- & 68.82 & -- & -- & 74.82 & -- & -- & 70.96 & -- & -- & \underline{75.75} & -- & -- & 39.34 & -- & -- \\
        WorldMirror & 68.55 & -- & -- & 67.81 & -- & -- & 68.28 & -- & -- & 69.50 & -- & -- & 67.40 & -- & -- & 74.25 & -- & -- & 44.26 & -- & -- \\
        \midrule
        Qwen3.5-9B & 48.42 & 48.29 & 43.82 & 48.29 & 45.55 & 41.78 & 52.69 & 52.15 & 49.46 & 45.74 & 48.58 & 42.20 & 50.14 & 43.84 & 47.12 & 47.01 & 52.10 & 42.81 & 45.90 & 54.10 & 29.51 \\
        Qwen3.5-27B & 63.68 & 44.34 & 62.76 & 65.07 & 37.33 & 65.75 & 65.05 & 51.61 & \underline{63.44} & 61.35 & 46.81 & 59.22 & 64.93 & 38.36 & 66.85 & 62.87 & 51.50 & 58.68 & 60.66 & 40.98 & 60.66 \\
        \ourmethod & \textbf{77.63} & \textbf{81.32} & \textbf{73.03} & \textbf{76.71} & 77.74 & \textbf{78.77} & \textbf{73.12} & \textbf{86.02} & \textbf{64.52} & \textbf{81.56} & \textbf{81.91} & \textbf{72.70} & \textbf{77.26} & 81.37 & \textbf{71.78} & \textbf{78.74} & \textbf{82.04} & \textbf{75.75} & \underline{73.77} & \textbf{77.05} & \underline{65.57} \\
        \bottomrule
    \end{tabular}%
    }
\end{table*}

\paragraph{Preference-data generation.}
We sample paired source images and captions from WorldPlay~\cite{worldplay} and combine each source condition with one of 224 curated 11-step camera trajectories. Table~\ref{tab:trajectory-library} summarizes the training trajectory library across nine sub-families. For each condition, we select two compatible world models from the pool in Section~\ref{sec:preference-data} and generate a matched video pair. Assignments are balanced across broadly compatible model pairs and restricted to the intersection of their supported action spaces, preventing unsupported controls from creating trivial preferences. We retain each model's default resolution to expose the reward model to varied spatial resolutions and aspect ratios. This process produces 50,000 video pairs at 15 fps using a fixed sampling seed of 1. After prepending an idle slot associated with the source frame, each 11-step trajectory contains 12 action slots, which we divide into three four-slot chunks. The resulting 150,000 candidate chunks are subsampled to approximately 100,000 examples for reasoning annotation and reward-model training. To mitigate candidate-position bias, we further augment samples from the majority preference combinations by swapping the two candidates and rewriting their annotations accordingly. Table~\ref{tab:position-augmentation} reports the training-data preference combinations before and after augmentation. We examine robustness to candidate ordering in Section~\ref{sec:position-bias-analysis}.

\paragraph{Benchmark generation.}
The generation procedure for \ourbench follows preference-data generation, but its generation conditions are held out from training. We use unseen source-image--caption pairs and independently construct a new library of 224 trajectories, each containing 23 action steps. As shown in Table~\ref{tab:trajectory-library}, the benchmark changes the distribution within the same trajectory taxonomy rather than reusing the training library. Each output contains 184 frames at 15 fps, yielding 760 matched video pairs. This separation tests whether a reward model has learned transferable action--video relations instead of memorizing particular prompts or trajectories.

\paragraph{Reward-model training.}
We initialize \ourmethod from Qwen3.5-9B~\cite{qwen35} and train it for 3 epochs with a global batch size of 128. We use a learning rate of $8\times10^{-6}$ with a cosine schedule and a warm-up ratio of 0.03. Sequence packing is enabled to reduce padding overhead.

\paragraph{Reward-model baselines and metric.}
We compare \ourmethod with four complementary categories of reward models and evaluators. Gemini~3.1 Pro~\cite{gemini31pro} and GPT-5.5~\cite{gpt55systemcard} serve as general-purpose proprietary VLM baselines. VideoAlign~\cite{videoalign}, UnifiedReward-Flex (UR-Flex)~\cite{unifiedrewardflex}, and UnifiedReward-Think (UR-Think)~\cite{unifiedreward_think} represent learned image and video preference models. LAION Aesthetic Predictor~\cite{laionaesthetics} and HPSv3~\cite{ma2025hpsv3} provide image-level quality signals, while DepthAnything3 (DAv3)~\cite{depthanything3} and WorldMirror~\cite{worldmirror} evaluate action consistency through geometric cues. We additionally evaluate the Qwen3.5-9B base model used to initialize \ourmethod and include the Qwen3.5-27B model as a larger-scale reference~\cite{qwen35}. Following Equation~\ref{eq:benchmark-accuracy}, we measure the agreement between each applicable method and human annotations under the three-way preference setting. For \ourmethod, Temporal Consistency and Artifacts \& Structure Integrity determine appearance, while Dynamic Generation Quality determines motion. If the two appearance criteria favor different videos within a chunk, its appearance prediction is Tie; chunk-level predictions are then aggregated by voting into video-level preferences.

\subsection{Reward-Model Evaluation on \ourbench}

\paragraph{Overall agreement with human preferences.}
Table~\ref{tab:benchmark-predictor-comparison} shows that \ourmethod achieves the highest overall agreement with human judgments across action consistency, appearance quality, and motion quality. Geometry estimators provide competitive action signals but cannot evaluate the visual consequences of the executed motion, whereas image and video preference models capture aspects of perceptual quality without assessing whether the commanded actions are followed. By jointly reasoning about action execution and the resulting visual content, \ourmethod provides reliable yet distinct judgments for all three dimensions within a single model. Notably, although \ourmethod is trained with supervision from both closed-source VLMs, it ultimately surpasses them across all three dimensions. We attribute this improvement to our annotation quality-control pipeline and examine its contribution in Section~\ref{sec:annotation-qc-ablation}.

\paragraph{Generalization across trajectories and visual domains.}
Across trajectory groups, \ourmethod achieves the highest action and motion agreement for translation, rotation, and compound controls, together with the highest appearance agreement on rotation and compound trajectories. The strong performance on compound trajectories is particularly notable, as these cases require the model to associate multiple consecutive controls with their corresponding visual changes, indicating that its action judgments remain reliable beyond isolated movements. Across visual domains, \ourmethod leads all three dimensions on game/anime content, achieves the highest action and motion agreement on photorealistic content, and attains the highest appearance agreement on artistic content, remaining competitive on the other dimensions. Together, these results indicate that \ourmethod generalizes reliably across varied trajectories and visual domains.

\subsection{Applied to World-Model Post-Training}

\paragraph{Controlled RL setup.}
To test whether preference accuracy translates into a useful learning signal, we post-train HY-WorldPlay~1.5~\cite{worldplay} under the WorldCompass protocol~\cite{worldcompass}, keeping the base model, training data, evaluation split, rollout procedure, and DiffusionNFT~\cite{diffusionnft} configuration unchanged and modifying only the reward signals. Training uses 4,000 diverse image--caption conditions paired with randomly constructed trajectories emphasizing compositions of the eight basic translation and rotation controls. Evaluation uses the same 600-case WorldPlay split~\cite{worldplay} and measures basic and combined actions at short ($\sim$125 frames), medium ($\sim$253 frames), and long ($\sim$381 frames) horizons, where DepthAnything3~\cite{depthanything3} estimates the camera trajectory of each generated video to compute action-adherence accuracy, and HPSv3~\cite{ma2025hpsv3} evaluates visual quality. This controlled setting isolates the effect of replacing heterogeneous geometry and image rewards with the two preferences produced by \ourmethod.

\begin{table*}[t]
    \centering
    \caption{\textbf{World-model post-training results on the WorldPlay split~\cite{worldplay}.} Action: action-adherence accuracy (\%) estimated by DepthAnything3~\cite{depthanything3}; Quality: HPSv3 score~\cite{ma2025hpsv3}. \ourmethod-RL differs from WorldCompass only in the reward signals. Best results are bold.}
    \label{tab:world-model-action-quality}
    \small
    \setlength{\tabcolsep}{7pt}
    \renewcommand{\arraystretch}{1.08}
    \begin{tabularx}{\textwidth}{@{}l>{\hsize=1.35\hsize\raggedright\arraybackslash}X*{4}{>{\hsize=0.9125\hsize\centering\arraybackslash}X}@{}}
        \toprule
        \multirow{2}{*}{\textbf{Horizon}}
        & \multirow{2}{*}{\textbf{Method}}
        & \multicolumn{2}{c}{\textbf{Combined Action}}
        & \multicolumn{2}{c}{\textbf{Basic Action}} \\
        \cmidrule(lr){3-4}\cmidrule(lr){5-6}
        & & \textbf{Action $\uparrow$} & \textbf{Quality $\uparrow$}
        & \textbf{Action $\uparrow$} & \textbf{Quality $\uparrow$} \\
        \midrule
        \multirow{3}{*}{\shortstack[l]{Short-term\\($\sim$125 frames)}}
        & HY-WorldPlay~1.5 & 21.74 & -1.05 & 62.33 & 1.96 \\
        & WorldCompass & 58.20 & 0.42 & 68.62 & 3.77 \\
        & \ourmethod-RL & \textbf{59.96} & \textbf{0.68} & \textbf{74.43} & \textbf{3.92} \\
        \midrule
        \multirow{3}{*}{\shortstack[l]{Mid-term\\($\sim$253 frames)}}
        & HY-WorldPlay~1.5 & 19.73 & -0.19 & 63.35 & 1.91 \\
        & WorldCompass & 55.01 & 0.37 & 74.09 & 3.61 \\
        & \ourmethod-RL & \textbf{57.79} & \textbf{0.55} & \textbf{77.27} & \textbf{3.85} \\
        \midrule
        \multirow{3}{*}{\shortstack[l]{Long-term\\($\sim$381 frames)}}
        & HY-WorldPlay~1.5 & 19.70 & -0.33 & 64.28 & 1.90 \\
        & WorldCompass & 54.82 & 0.73 & 76.56 & 3.72 \\
        & \ourmethod-RL & \textbf{56.40} & \textbf{1.02} & \textbf{78.84} & \textbf{3.94} \\
        \bottomrule
    \end{tabularx}
\end{table*}

\paragraph{Quantitative comparison.}
Table~\ref{tab:world-model-action-quality} shows how the choice of reward affects the post-trained world model. Compared with WorldCompass~\cite{worldcompass}, \ourmethod improves combined-action accuracy by 1.58--2.78 points and basic-action accuracy by 2.28--5.81 points across the three generation horizons. The consistent gains on combined actions are particularly encouraging, since errors are more likely to accumulate when several controls must be executed in sequence. Visual quality improves at the same time: HPSv3~\cite{ma2025hpsv3} increases in all six settings, by 0.18--0.29 for combined actions and 0.15--0.24 for basic actions. At the longest horizon, for example, \ourmethod reaches HPSv3 scores of 1.02 and 3.94, compared with 0.73 and 3.72 for WorldCompass. Thus, the improvement in command following is maintained as the generation becomes longer and does not come at the cost of visual quality.

\begin{figure*}[t]
    \centering
    \includegraphics[width=0.94\textwidth]{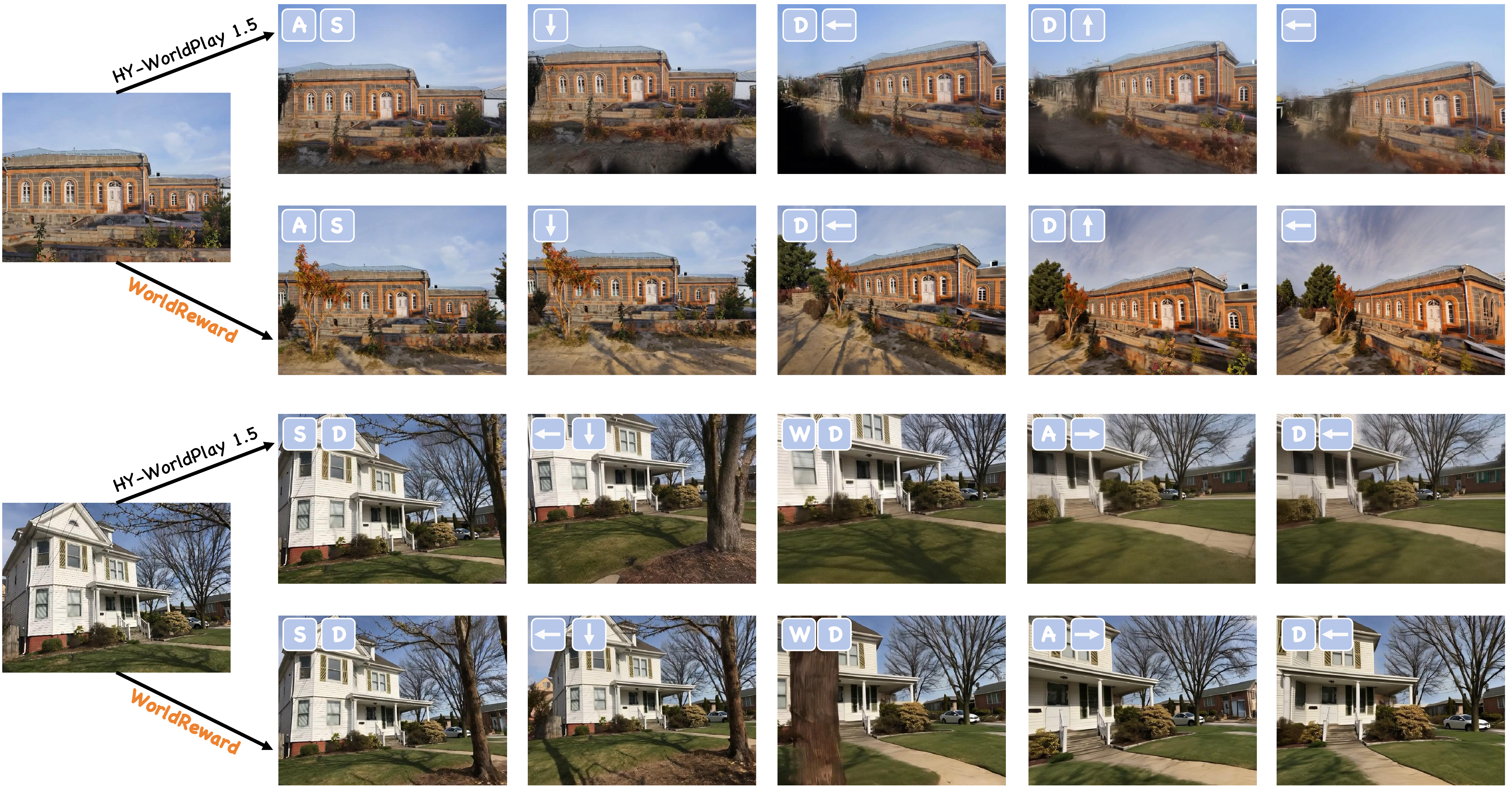}
    \caption{\textbf{Qualitative comparison with HY-WorldPlay~1.5~\cite{worldplay}.} Top: HY-WorldPlay~1.5; bottom: \ourmethod-RL (labeled WorldReward). Icons mark the translation key and rotation direction active at each frame.}
    \label{fig:qualitative-vs-worldplay}
\end{figure*}

\paragraph{Qualitative comparison with HY-WorldPlay~1.5.}
Figure~\ref{fig:qualitative-vs-worldplay} shows the effect of post-training on both action following and visual quality. HY-WorldPlay~1.5~\cite{worldplay} responds weakly to several controls, producing limited viewpoint changes over successive steps, and its outputs degrade as the trajectory proceeds, with smeared regions and blurred scene structure emerging in later frames. In contrast, our post-trained model executes the requested translations and rotations more distinctly while keeping the scene structure clean and coherent throughout the trajectory. The improvement is visible in both examples and is consistent with the action-accuracy and visual-quality gains in Table~\ref{tab:world-model-action-quality}.

\begin{figure*}[t]
    \centering
    \includegraphics[width=0.94\textwidth]{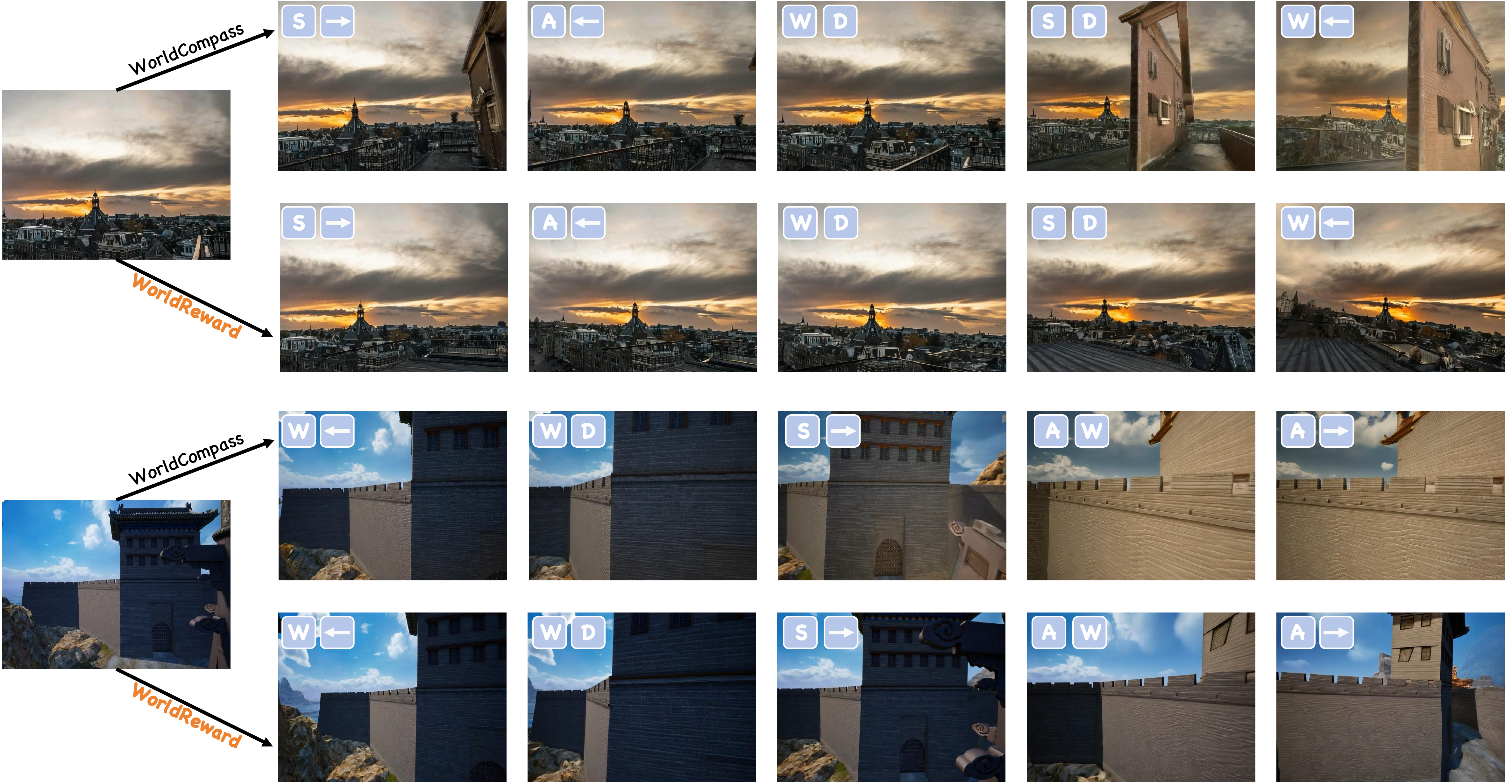}
    \caption{\textbf{Qualitative comparison with WorldCompass~\cite{worldcompass}.} Top: WorldCompass; bottom: \ourmethod-RL (labeled WorldReward). Layout as in Figure~\ref{fig:qualitative-vs-worldplay}.}
    \label{fig:qualitative-vs-worldcompass}
\end{figure*}

\paragraph{Qualitative comparison with WorldCompass.}
Figure~\ref{fig:qualitative-vs-worldcompass} compares WorldCompass~\cite{worldcompass} with our post-trained model.
Relative to the base HY-WorldPlay~1.5 model~\cite{worldplay}, WorldCompass improves camera-control accuracy, but this gain is accompanied by pronounced color shifts in later frames and a gradual loss of texture consistency. One possible limitation is its image-based reward model, which scores sampled frames independently without directly assessing whether color and texture remain consistent over time. By evaluating visual quality from action-aligned video evidence, our post-trained model maintains a more stable appearance throughout the trajectory while retaining the improved action following. This qualitative difference complements the results in Table~\ref{tab:world-model-action-quality} and motivates temporal visual-quality feedback during post-training.

\begin{figure*}[t]
    \centering
    \includegraphics[width=\textwidth]{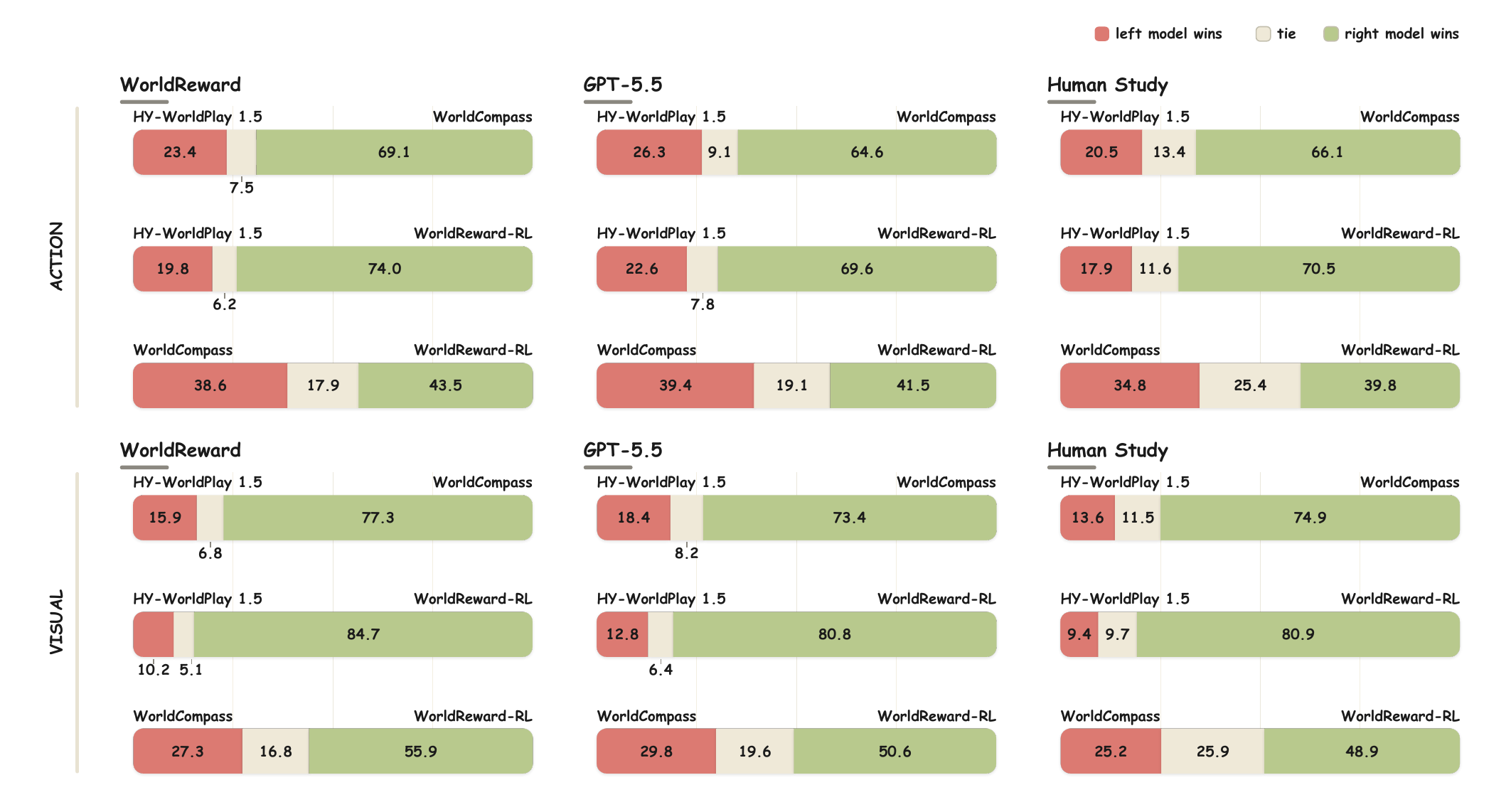}
    \caption{\textbf{Pairwise evaluation of post-trained world models.} Left-win / tie / right-win rates (\%) judged by \ourmethod, GPT-5.5~\cite{gpt55systemcard}, and human annotators for action consistency (top) and visual quality (bottom) on the 200 human-study pairs. \ourmethod-RL denotes HY-WorldPlay~1.5 post-trained with \ourmethod.}
    \label{fig:world-model-h2h}
\end{figure*}

\paragraph{Human evaluation protocol.}
We conduct a blinded study on 200 video pairs sampled from the long-horizon, combined-action subset, where differences in action execution and visual stability are most apparent. The two videos in each pair share the same source image, text prompt, and action trajectory. Their model identities are hidden and their left--right order is randomized. Each pair is independently judged by three annotators for action consistency and visual quality. For each dimension, annotators choose the left video, the right video, or Tie when neither video has a clear and consistently observable advantage. We report preferences over all individual judgments rather than applying majority voting, preserving genuine ambiguity between closely matched outputs. This yields 600 human judgments per dimension.

\paragraph{Pairwise preference evaluation.}
We compare the generated videos directly using WorldReward, GPT-5.5~\cite{gpt55systemcard}, and human evaluators under the same pairwise protocol, where each comparison selects one video or Tie (Figure~\ref{fig:world-model-h2h}). All three favor our post-trained model over the original HY-WorldPlay~1.5~\cite{worldplay} in both action consistency and visual quality. The same pattern holds against WorldCompass~\cite{worldcompass}. For action consistency, our post-trained model is preferred over WorldCompass by 43.5\% versus 38.6\% under WorldReward, 41.5\% versus 39.4\% under GPT-5.5, and 39.8\% versus 34.8\% in the human study. The difference is larger for visual quality: the corresponding preferences are 55.9\% versus 27.3\%, 50.6\% versus 29.8\%, and 48.9\% versus 25.2\%. Across both comparison pairs and evaluation dimensions, WorldReward, GPT-5.5, and human evaluators give the same overall ranking. These consistent aggregate trends indicate that the post-training gains are not specific to the reward used for optimization.

\subsection{Ablation and Discussion}

\paragraph{Impact of annotation refinement.}
\label{sec:annotation-qc-ablation}

\begin{table}[t]
    \centering
    \caption{\textbf{Annotation-refinement ablation on \ourbench (\%).} Rows 2--4 train the reward model on annotations after each pipeline stage; row 1 uses Gemini~3.1 Pro directly as the judge.}
    \label{tab:ablation-annotation-qc}
    \small
    \setlength{\tabcolsep}{5pt}
    \begin{tabularx}{\linewidth}{@{}>{\hsize=1.6\hsize\raggedright\arraybackslash}X*{4}{>{\hsize=0.85\hsize\centering\arraybackslash}X}@{}}
        \toprule
        \textbf{Setting} & \makecell{\textbf{Action}\\\textbf{Consistency}} & \makecell{\textbf{Appearance}\\\textbf{Quality}} & \makecell{\textbf{Motion}\\\textbf{Quality}} & \textbf{Average} \\
        \midrule
        Gemini 3.1 Pro (direct judge) & 65.79 & 80.13 & 60.79 & 68.90 \\
        \midrule
        Gemini distillation & 66.84 & 78.92 & 60.31 & 68.69 \\
        + Agent-harness QC & \underline{75.84} & \underline{80.46} & \underline{71.52} & \underline{75.94} \\
        + Human review & \textbf{77.63} & \textbf{81.32} & \textbf{73.03} & \textbf{77.33} \\
        \bottomrule
    \end{tabularx}
\end{table}

\begin{table}[t]
    \centering
    \caption{\textbf{Structured-evidence and reasoning-supervision ablations on \ourbench (\%).} Top: removing one visual component at both training and inference. Bottom: progressively adding reasoning targets to label-only supervision.}
    \label{tab:ablation-model-design}
    \small
    \setlength{\tabcolsep}{5pt}
    \begin{tabularx}{\linewidth}{@{}>{\hsize=1.6\hsize\raggedright\arraybackslash}X*{4}{>{\hsize=0.85\hsize\centering\arraybackslash}X}@{}}
        \toprule
        \textbf{Setting} & \makecell{\textbf{Action}\\\textbf{Consistency}} & \makecell{\textbf{Appearance}\\\textbf{Quality}} & \makecell{\textbf{Motion}\\\textbf{Quality}} & \textbf{Average} \\
        \midrule
        \multicolumn{5}{@{}l}{\textit{Structured visual evidence}} \\
        w/o source image & 76.84 & 78.95 & 72.41 & 76.07 \\
        w/o frame-grid overview & 74.92 & 79.88 & 69.74 & 74.85 \\
        w/o action-level panels & 74.81 & 80.56 & 71.68 & 75.68 \\
        \addlinespace[2pt]
        \multicolumn{5}{@{}l}{\textit{Reasoning supervision}} \\
        Preference labels only & 71.84 & 77.96 & 67.61 & 72.47 \\
        + Overall comparison summary & 74.93 & 79.52 & 70.38 & 74.94 \\
        + Per-video analysis (full model) & \textbf{77.63} & \textbf{81.32} & \textbf{73.03} & \textbf{77.33} \\
        \bottomrule
    \end{tabularx}
\end{table}

\begin{table*}[t]
    \centering
    \begin{minipage}[t]{0.48\textwidth}
        \centering
        \caption{\textbf{Position-swap augmentation.} Training-sample counts per preference combination; the first letter is the action winner and the second the visual-quality winner (e.g., AB: A wins action, B wins visual quality).}
        \label{tab:position-augmentation}
        \small
        \setlength{\tabcolsep}{2pt}
        \renewcommand{\arraystretch}{1.01}
        \begin{tabularx}{\linewidth}{@{}>{\centering\arraybackslash}X*{2}{>{\centering\arraybackslash}X}@{}}
            \toprule
            \textbf{Combo} & \textbf{Original} & \textbf{Augmented} \\
            \midrule
            AA & 37,107 & 37,107 \\
            AB & 25,365 & 25,365 \\
            BA & 17,318 & 25,364 \\
            BB & 20,164 & 37,106 \\
            \midrule
            Total & 99,954 & 124,942 \\
            \bottomrule
        \end{tabularx}
        \vspace{3pt}
    \end{minipage}
    \hfill
    \begin{minipage}[t]{0.48\textwidth}
        \centering
        \caption{\textbf{Robustness to candidate ordering on \ourbench (\%).} Randomized: the A/B order of every chunk is shuffled independently and predictions are mapped back before voting.}
        \label{tab:position-robustness}
        \footnotesize
        \setlength{\tabcolsep}{1pt}
        \renewcommand{\arraystretch}{1.12}
        \begin{tabularx}{\linewidth}{@{}>{\hsize=1.45\hsize\raggedright\arraybackslash}X*{3}{>{\hsize=0.85\hsize\centering\arraybackslash}X}@{}}
            \toprule
            \textbf{Setting} & \textbf{Action} & \textbf{Appearance} & \textbf{Motion} \\
            \midrule
            Original order & 77.63 & 81.32 & 73.03 \\
            Randomized order & 76.39 & 80.79 & 72.08 \\
            $\Delta$ & $-1.24$ & $-0.53$ & $-0.95$ \\
            \bottomrule
        \end{tabularx}
    \end{minipage}
\end{table*}

\begin{table*}[!t]
    \centering
    \begin{minipage}[t]{0.48\textwidth}
        \centering
        \caption{\textbf{Reward-component ablation} on long-horizon combined actions. Metrics as in Table~\ref{tab:world-model-action-quality}.}
        \label{tab:reward-component-ablation}
        \small
        \setlength{\tabcolsep}{4pt}
        \begin{tabularx}{\linewidth}{@{}>{\raggedright\arraybackslash}X*{2}{>{\centering\arraybackslash}p{0.21\linewidth}}@{}}
            \toprule
            \textbf{Reward} & \textbf{Action $\uparrow$} & \textbf{Quality $\uparrow$} \\
            \midrule
            None (HY-WorldPlay~1.5) & 19.70 & -0.33 \\
            Action only & 52.73 & 0.21 \\
            Visual only & 27.82 & 0.89 \\
            Action + Visual & \textbf{56.40} & \textbf{1.02} \\
            \bottomrule
        \end{tabularx}
    \end{minipage}
    \hfill
    \begin{minipage}[t]{0.48\textwidth}
        \centering
        \caption{\textbf{Pair-level agreement with human judgments (\%).} Computed over all 600 individual judgments per dimension; parentheses give pair-clustered 95\% bootstrap CIs.}
        \label{tab:self-evaluation-agreement}
        \small
        \setlength{\tabcolsep}{3pt}
        \begin{tabularx}{\linewidth}{@{}>{\raggedright\arraybackslash}X*{2}{>{\centering\arraybackslash}p{0.34\linewidth}}@{}}
            \toprule
            \textbf{Evaluator} & \textbf{Action} & \textbf{Visual Quality} \\
            \midrule
            GPT-5.5 & 69.1 {\scriptsize (65.6--72.5)} & 72.0 {\scriptsize (68.7--75.2)} \\
            WorldReward & \textbf{70.3} {\scriptsize (66.9--73.6)} & \textbf{72.9} {\scriptsize (69.6--76.0)} \\
            \bottomrule
        \end{tabularx}
    \end{minipage}
\end{table*}

Table~\ref{tab:ablation-annotation-qc} shows that distillation from Gemini~3.1 Pro~\cite{gemini31pro} largely preserves the direct judge's average agreement, with a modest gain in action consistency but small decreases in appearance and motion quality. The substantial improvement emerges only after agent-harness quality control with GPT-5.5~\cite{gpt55systemcard}, which raises the average agreement from 68.69\% to 75.94\%. This gain is concentrated in action consistency and motion quality, where revisiting local action--video evidence helps correct errors inherited from the initial distilled annotations. Human review provides a further 1.39-point improvement and produces the best agreement across all three dimensions. These results indicate that the final advantage over direct VLM judging comes primarily from annotation refinement rather than distillation alone.

\paragraph{Contribution of structured visual evidence.}
\label{sec:structured-visual-ablation}
We train three variants, each removing one visual component during both training and inference, while keeping the initialization, data, supervision, and optimization settings fixed. As shown in the first block of Table~\ref{tab:ablation-model-design}, removing the source image mainly affects appearance quality ($-2.37$ points), since the model loses the reference needed to identify drift in objects, layout, and scene content. The temporal views have distinct effects. Removing the frame-grid overview causes the largest average drop ($-2.48$ points), including decreases of 2.71 and 3.29 points in action and motion agreement. Removing the action-level panels mainly reduces action agreement ($-2.82$ points), with a smaller effect on motion ($-1.35$ points). The frame grid captures the evolution of the chunk, whereas the panels directly associate each control with its local visual change. Their complementary roles explain why the full input performs best across all three dimensions.

\paragraph{Role of structured reasoning supervision.}
\label{sec:reasoning-supervision-ablation}
We ablate the reasoning targets while keeping the training examples and final preference labels fixed. As shown in the second block of Table~\ref{tab:ablation-model-design}, the label-only model reaches 72.47\% average agreement. Adding an overall comparison between the two video chunks improves the average to 74.94\%, with larger gains on action consistency and motion quality than on appearance quality. Adding per-video analysis further raises the average by 2.39 points and yields the best result on all three dimensions, with the largest additional gains in action consistency ($+2.70$) and motion quality ($+2.65$). Unlike the overall summary, this supervision requires the model to inspect each chunk separately before making the comparison, making local action failures or motion artifacts less likely to be hidden by a brief pair-level judgment.

\paragraph{Robustness to candidate ordering.}
\label{sec:position-bias-analysis}
To test whether the model relies on candidate position, we independently randomize the A/B order of every chunk in \ourbench and map the predictions back to the original candidate identities before video-level aggregation. As shown in Table~\ref{tab:position-robustness}, randomized ordering reduces agreement by 1.24, 0.53, and 0.95 points for action consistency, appearance quality, and motion quality, respectively. The limited degradation indicates that \ourmethod is largely robust to candidate ordering, suggesting that the swap-based augmentation applied during training data construction (Table~\ref{tab:position-augmentation}) effectively mitigates candidate-position bias.

\paragraph{Contributions of action and visual rewards.}
\label{sec:reward-component-ablation}
We isolate the two preferences used for post-training by optimizing HY-WorldPlay~1.5~\cite{worldplay} with the action reward alone, the visual reward alone, or their combination. We report results on long-horizon combined actions, where failures in control and visual coherence are most likely to accumulate. Table~\ref{tab:reward-component-ablation} shows that the two rewards serve different roles. Action-only optimization substantially improves action accuracy, but its weaker visual quality can make the executed camera motion less reliably recognizable, leaving its action score below that of the combined reward. Visual-only optimization raises HPSv3~\cite{ma2025hpsv3} more clearly while providing limited guidance for executing a sequence of controls. Combining the two achieves the best result on both measures: the action reward encourages trajectory adherence, while the visual reward helps preserve the appearance and temporal coherence needed for that motion to remain identifiable over long horizons.

\paragraph{Pair-level agreement with human judgments.}
\label{sec:self-evaluation-analysis}
The aggregate results above show that WorldReward and human evaluators favor the same model, but they do not measure whether the two make the same decision on individual video pairs. We therefore compare the predictions of WorldReward and GPT-5.5~\cite{gpt55systemcard} with the independent annotations collected in the human study. Agreement is computed against all three annotations for each pair. We obtain 95\% confidence intervals by bootstrapping video pairs, so that annotations of the same pair are not treated as independent samples. Table~\ref{tab:self-evaluation-agreement} shows that WorldReward reaches 70.3\% agreement with human judgments on action consistency and 72.9\% on visual quality, comparable to GPT-5.5 on both dimensions. The result complements the aggregate preferences in Figure~\ref{fig:world-model-h2h}: WorldReward not only favors the same model as human evaluators overall, but also makes similar judgments on individual pairs. This pair-level correspondence provides further evidence that the reward remains aligned with human assessments when it is used to evaluate the model it optimized.

\FloatBarrier

\section{Conclusion}

We presented \ourmethod, a VLM-based pairwise reward model that jointly evaluates action consistency and visual quality for camera-conditioned world models. \ourmethod decomposes long videos into action-aligned chunks, reasons over structured visual evidence, and aggregates chunk-level decisions into video-level preferences. To train it, we construct a large-scale reasoning-augmented preference dataset generated from a frontier VLM and refined through agent auditing and human review. We further introduce \ourbench for measuring agreement with human preferences across action, appearance, and motion, on which \ourmethod achieves the highest agreement on all three dimensions. When used for RL post-training of HY-WorldPlay~1.5, \ourmethod improves both action execution and visual quality across generation horizons.


\bibliographystyle{unsrtnat}
\bibliography{reference}



\end{document}